\documentclass[acmtog]{acmart}

\usepackage{booktabs} 

\usepackage[ruled]{algorithm2e} 

\SetAlFnt{\small}
\SetAlCapFnt{\small}
\SetAlCapNameFnt{\small}
\SetAlCapHSkip{0pt}

\copyrightyear{2024}
\acmYear{2024}
\setcopyright{rightsretained}
\acmConference[SA Conference Papers '24]{SIGGRAPH Asia 2024 Conference Papers}{December 3--6, 2024}{Tokyo, Japan}
\acmBooktitle{SIGGRAPH Asia 2024 Conference Papers (SA Conference Papers '24), December 3--6, 2024, Tokyo, Japan}

\received[final version]{September 2024}
\received[accepted]{August 2024}

\begin{teaserfigure}
    \centering
    \includegraphics[width=\textwidth]{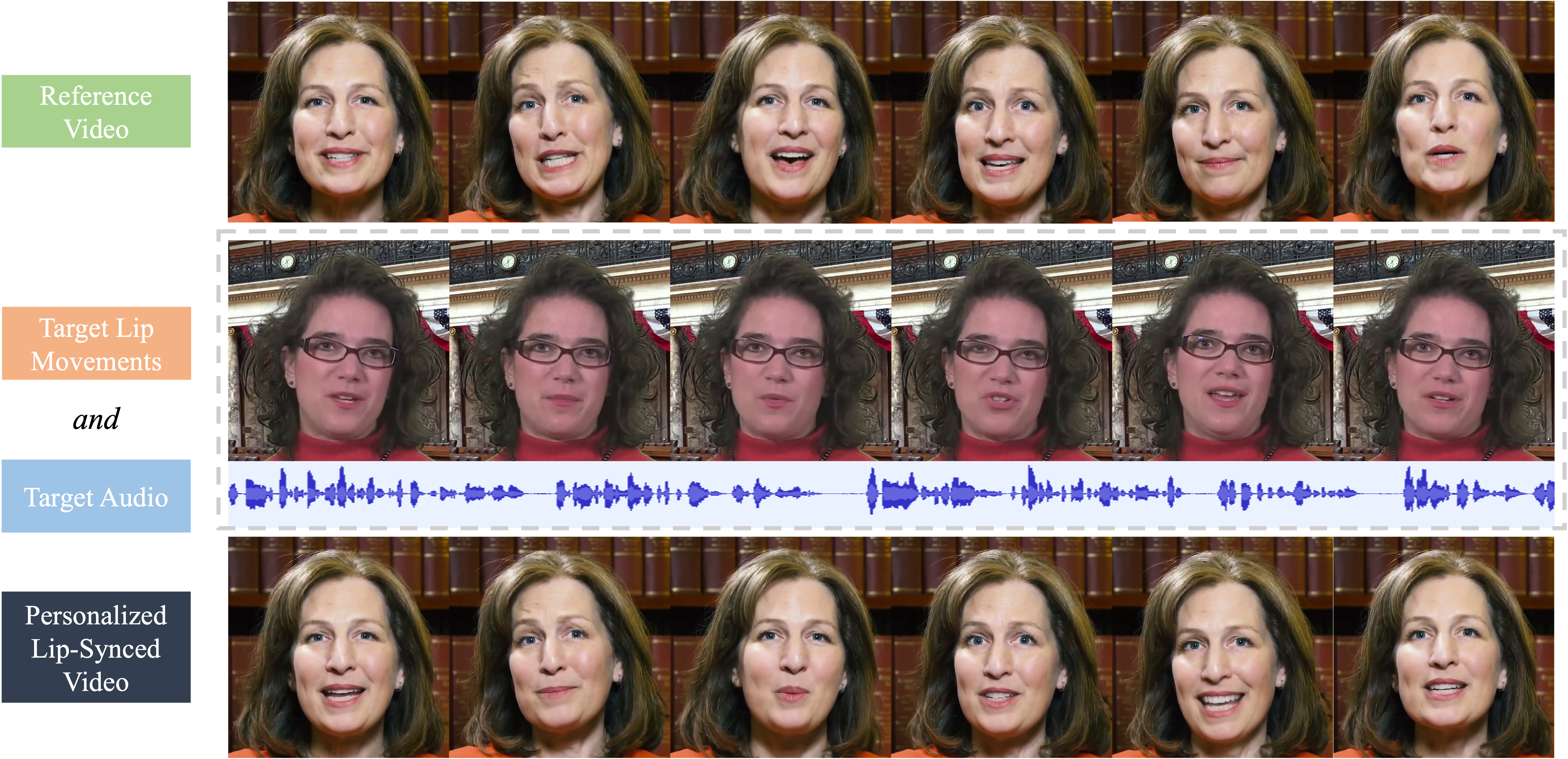}
    \caption{Given an arbitrary input (reference) video and a target audio, our method can synthesize a high-fidelity and personalized lip-synced video while highlighting speaker's \emph{persona}, like speaking style and facial details.}
    \label{fig:teaser}
\end{teaserfigure}

\begin{document}
\title{PersonaTalk: Bring Attention to Your Persona in Visual Dubbing}

\author{Longhao Zhang}
\affiliation{%
 \institution{Bytedance}
 \country{China}
}
\email{zhanglonghao.zlh@bytedance.com}
\authornote{Authors contributed equally to this research work.}

\author{Shuang Liang}
\affiliation{%
 \institution{Bytedance}
 \country{China}
}
\email{liangshuang.echo@bytedance.com}
\authornotemark[1]

\author{Zhipeng Ge}
\affiliation{%
 \institution{Bytedance}
 \country{China}
}
\email{zhipengge@bytedance.com}
\authornotemark[1]

\author{Tianshu Hu}
\affiliation{%
 \institution{Bytedance}
 \country{China}
}
\email{tianshu.hu@bytedance.com}
\authornotemark[1]
\authornote{Corresponding author.}

\renewcommand\shortauthors{Longhao Zhang, Shuang Liang, Zhipeng Ge and Tianshu Hu}

\begin{abstract}
For audio-driven visual dubbing, it remains a considerable challenge to uphold and highlight speaker's persona while synthesizing accurate lip synchronization. Existing methods fall short of capturing speaker's unique speaking style or preserving facial details. In this paper, we present $PersonaTalk$, an attention-based two-stage framework, including geometry construction and face rendering, for high-fidelity and personalized visual dubbing. In the first stage, we propose a style-aware audio encoding module that injects speaking style into audio features through a cross-attention layer. The stylized audio features are then used to drive speaker's template geometry to obtain lip-synced geometries. In the second stage, a dual-attention face renderer is introduced to render textures for the target geometries. It consists of two parallel cross-attention layers, namely Lip-Attention and Face-Attention, which respectively sample textures from different reference frames to render the entire face. With our innovative design, intricate facial details can be well preserved.
Comprehensive experiments and user studies demonstrate our advantages over other state-of-the-art methods in terms of visual quality, lip-sync accuracy and persona preservation. Furthermore, as a person-generic framework, $PersonaTalk$ can achieve competitive performance as state-of-the-art person-specific methods. Project Page: \href{https://grisoon.github.io/PersonaTalk/}{\emph{https://grisoon.github.io/PersonaTalk/}}.

\end{abstract}

%
%


%
%


\maketitle

\section{introduction}

Audio-driven visual dubbing broadly extends its application in real-world scenarios, such as creating digital human oral broadcast, translating human videos to various languages or altering recorded video contents.
Existing methods are mainly implemented in two different ways. Some works \cite{SynthesizingObama, LSP, AD-NeRF, RAD-NeRF, ER-NeRF, GeneFace, SyncTalk} can generate high-fidelity results through person-specific training or fine-tuning, yet require plenty of target speaker's videos for personalized modeling. In addition, the customized training process also limits their application and popularization. Other works \cite{kr2019towards, Wav2Lip, MakeItTalk, doukas2021headgan, PC-AVS, biswas2021realistic, videoretalking, StyleSync, TalkLip, IP-LAP, DINet, DiffusedHead, DreamTalk} explore to train a universal model.
Despite recent advances in synthesizing lip synchronization, these person-generic methods fail to preserve speaker's persona, like speaking style and facial details. In particular, a number of them \cite{PC-AVS, videoretalking} achieve audio-lip sync through the AdaIN layers \cite{adain}, which directly interpose audio features to the renderer and implicitly edit lips. We argue that this leads to imprecise and homogenized lip movements. While IP\_LAP \cite{IP-LAP} obtains the 2D landmarks through audio features, and explicitly warps the face with these lip-synced landmarks, yet it ignores the speaking style. Besides, averaging the warped features or images causes blurriness and loss of facial details.

In this paper, we bring $attention$ to speaker's persona, proposing an attention-based two-stage framework $PersonaTalk$, which is capable of not only producing lip movements that are well synchronized with the audio, but also upholding speaker's unique speaking style and facial details.
Specifically, in the first stage, we start by extracting video speaker's 3D facial geometric information using a carefully designed hybrid geometry estimation approach. We hypothesize that introducing 3D geometry \cite{3DMM, Deep3D, DECA, gecer2019ganfit} as the intermediate representation is crucial, because a) learning "audio-to-geometry" is much easier than learning "audio-to-image" since the former does not involved head pose or texture, which produces more precise lip movements; b) speaking style can be learned from the geometric statistical characteristics, which is inspired by \cite{Audio2Face, VOCA, MeshTalk, FaceFormer, CodeTalker, Imitator} and described in detail in Section \ref{sec:sagc}. Then we inject the style into the audio features by a single cross-attention layer, and employ it to drive the template geometry to obtain the target lip-synced geometries through multiple cross-attention and self-attention layers. Taking the lip-synced geometries as priors, in the second stage, we generate the target talking face with a novel dual-attention face renderer.
It consists of two parallel cross-attention layers: Lip-Attention and Face-Attention. The former samples lip-related textures from lip-reference frames, while the latter samples other facial textures from face-reference frames. Furthermore, to synthesize realistic faces while preserving speaker's facial details (e.g., Teeth, face outline, skin tone, makeup, lighting, etc.) as much as possible, we design an elaborate strategy to select different reference frames for them. Detailed description is in Section \ref{sec:dafr}. 

Our main contributions can be summarized as:

1) We propose an attention-based two-stage framework for high-fidelity and persona-preserved visual dubbing.

2) We introduce 3D facial geometry as the intermediate representation and extract speaker's speaking style from it. The style embedding is then injected into audio features to achieve stylized driving.

3) We devise a novel dual-attention face renderer, which consists of Lip-Attention and Face-Attention. It can generate photo-realistic images with intricate facial details.

4) Extensive experiments and user studies demonstrate our advantages over other state-of-the-art methods. Moreover, as a person-generic framework, our approach can achieve competitive performance as person-specific methods.

\section{related work}

\noindent\textbf{Person-Generic Visual Dubbing}. 
Person-generic methods\cite{10.1145/3449063, shen2023difftalk} intend to build a universal model that can be generalized to any speaker without retraining or fine-tuning. 
LipGAN \cite{kr2019towards} introduces a GAN-based network to inpaint the masked lower face. 
Wav2Lip \cite{Wav2Lip} further proposes a SyncNet\cite{chung2017out} as the lip-sync discriminator to achieve better audio-visual alignment. 
Inspired by 3D pose prior in talking faces, PC-AVS \cite{PC-AVS} modularizes the representations of talking faces into the spaces of speech content, head pose, and identity respectively through a compact pose code. 
DINet \cite{DINet} modularizes this task into deformation and inpainting stages to facilitate facial dubbing in high-resolution video content. 
Further advancements have been made by VideoReTalking \cite{videoretalking}, which develops an expression editing network to eliminate expression, a lip-sync network to generate target lip movements, and an identity-aware enhancement network to improve photo-realism. 
TalkLip \cite{TalkLip} escalates lip-speech synchronization via contrastive learning coupled with a transformer-based audio encoder. 
StyleSync \cite{StyleSync} innovates a StyleGAN-based generator \cite{stylegan} with the mask-based spatial information encoding to obtain high-fidelity results on a variety of scenes.
Recently, diffusion models \cite{ho2020denoising} have risen to prominence for their effectiveness in various generative tasks.
DiffusedHead \cite{DiffusedHead} produces top-tier results adaptable to a range of speaking styles. However, it relies on videos with static backgrounds for training, which presents challenges when applied to extensive in-the-wild datasets.
DiffTalk \cite{shen2023difftalk} attempts to extend model generalization, but struggles with significant temporal flickering and poor lip accuracy in cross-identity results.
It is worth noting that some works like \cite{10.1145/3474085.3475280} and \cite{styletalk} also explore how to extract the speaking style. However, our work is different from theirs. We learn style from geometric statistical characteristics and inject them into audio features through cross-attention layers.

{\noindent\textbf{Person-Specific Visual Dubbing}. 
Person-specific visual dubbing \cite{huang2024makeyouranchor} is much easier than the generic one, since they are limited to the certain person in the known environment.
SynthesizingObama \cite{SynthesizingObama} collects 17 hours of training data to achieve high-quality dubbing.
With the development of Neural Radiance Fields (NeRF) \cite{nerf}, numerous works adapt NeRF in the field of lip sync \cite{AD-NeRF, RAD-NeRF, ER-NeRF, SyncTalk}. 
Typically, RAD-NeRF \cite{RAD-NeRF} enables real-time talking portraits synthesizing, making this kind of method more practical. 
ER-NeRF \cite{ER-NeRF} uses a cross-modal attention mechanism to obtain more suitable features to model accurate facial motion.  
Although these methods have made great progress in terms of efficiency and realism, they are only capable of generating videos of specific person, meanwhile, the training process or the fine-tuning process is time-consuming.
 }
\section{method}

\begin{figure*}
  \centering
  \includegraphics[width=0.9\linewidth]{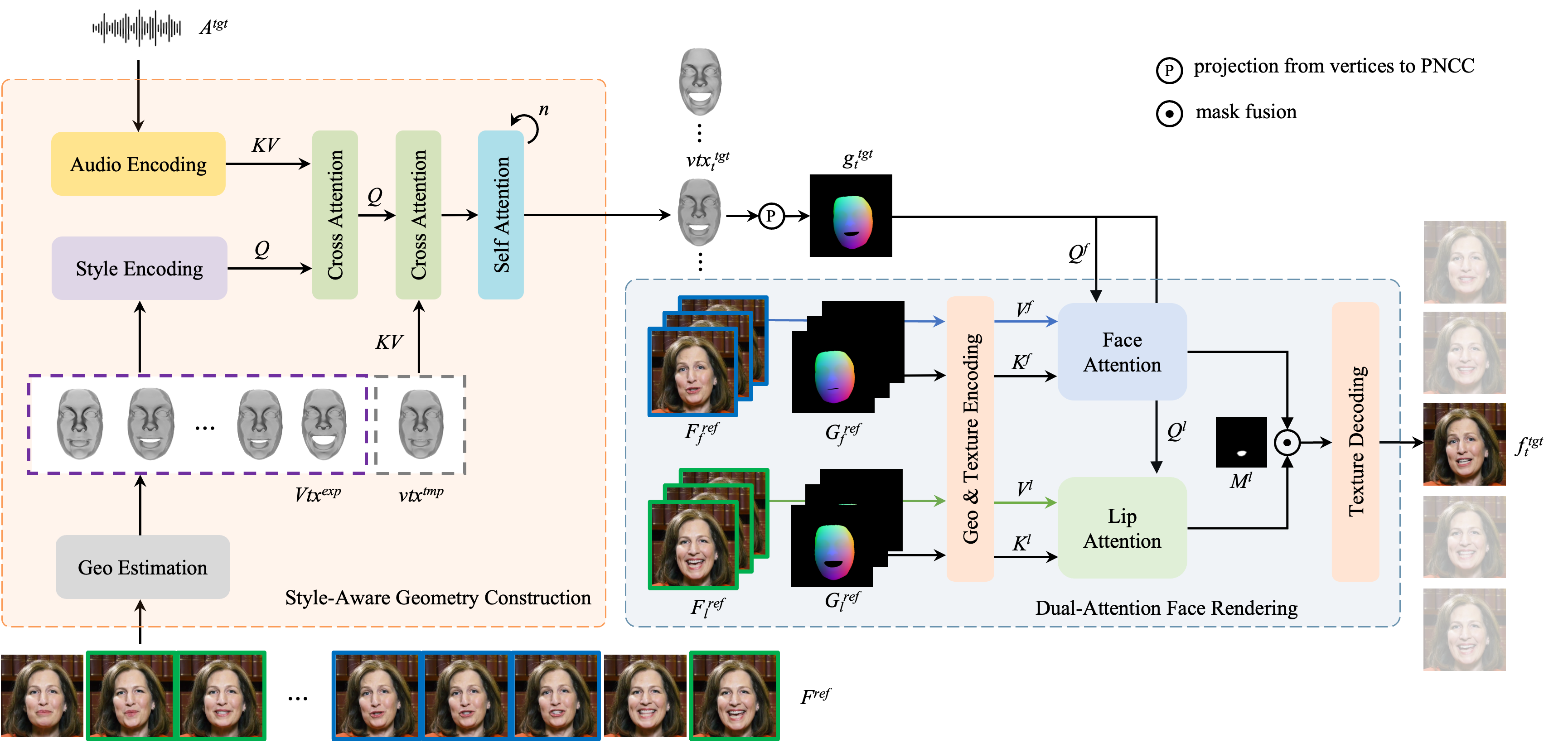}
  \caption{Our proposed method is an attention-based two-stage framework consisting of geometry construction and face rendering. Note that the frames in the green box are selected as the lip-reference, while the frames in the blue box are selected as the face-reference.}
  \label{fig:one}
\end{figure*}

We propose an attention-based two-stage framework consisting of geometry construction and face rendering. The former estimates the speaker's facial geometry and speaking style from the reference video $F^{ref}_{1:T}$ and constructs the target geometries $G^{tgt}_{1:T}$ which are lip-synced with the target audio $A^{tgt}_{1:T}$ while retaining the personalized speaking style. The latter renders the target talking faces $F^{tgt}_{1:T}$ with the geometries generated in the first stage and textures sampled from the reference video. Owing to our novel dual-attention face renderer, facial details can be effectively retained during rendering. For ease of writing, we assume that the reference video and target audio have the same length $T$.

\subsection{Style-Aware Geometry Construction}\label{sec:sagc}
In this stage, we first employ a hybrid geometry estimation approach with our carefully designed loss functions to obtain speaker's facial geometric coefficients $C_{1:T}$ from $F^{ref}_{1:T}$. Then we extract and encode the audio features from $A^{tgt}_{1:T}$. Note that we learn personalized speaking style $S$ from the geometric statistical characteristics and inject it to the audio features, making this step style-aware. Finally, we generate $G^{tgt}_{1:T}$ from $C_{1:T}$ and $A^{tgt}_{1:T}$.

\noindent\textbf{Facial Geometry Estimation.} 
We adopt the hybrid geometry estimation paradigm, including learning-based initialization and optimization-based tuning, to extract video speaker's 3DMM coefficients. Specifically, given the reference video frames $F^{ref}_{1:T}$, we first use an off-the-shelf network \cite{Deep3D} to predict the initial coefficients ${C}^{init}_{1:T}=\{\{{\alpha}^{init}_{1}, {\beta}^{init}_{1}, {p}^{init}_{1}\}, \dots, \{{\alpha}^{init}_{T}, {\beta}^{init}_{T}, {p}^{init}_{T}\}\}$, where $\alpha_{t}$, $\beta_{t}$ and $p_{t}$ denote shape, expression and pose, respectively. Subsequently, we 
iteratively optimize ${C}_{1:T}$ following \cite{gecer2019ganfit}. On the basis of it, we introduce a temporal-smooth loss $L_{tempo}$ to ensure temporal stability and accuracy, and a regularization loss $L_{reg}$ to prevent entanglement. The former is written as  
\begin{equation}
\begin{aligned}
    L_{tempo} = &\sum_{k=1}^{K}\sum_{t=1}^{T-1} || (kp_{k,t+1} - kp_{k,t}) - (\overline{kp}_{k,t+1} - \overline{kp}_{k,t}) ||_2\\
    &+\lambda(||\nabla{p}_{t}||_2+||\nabla{\beta}_{t}||_2),
\end{aligned}
\end{equation}
where $KP_{1:K} = \{kp_1, \dots, kp_K\}$ are some selected mesh vertices projected onto the image plane, and $\overline{KP}_{1:K} = \{\overline{kp}_1, \dots, \overline{kp}_K\}$ are their corresponding ground-truth landmarks .$\nabla$ is the Laplace operator, and $\lambda$ is set to 0.2. The latter loss is a regularization term,
\begin{equation}
    L_{reg} = \sum_{t=1}^{T} (|| \alpha_{t} - \overline{\alpha}||_2 + 
                              || \beta_{t} - {\beta}^{init}_{t}||_2 + 
                              || p_{t} - {p}^{init}_{t}||_2),
\end{equation}
where $\overline{\alpha}$ is the initial mean shape coefficient over $T$.

\noindent\textbf{Style-Aware Audio Encoding.}
HuBERT \cite{HuBERT} has been proven to be able to transform the raw audio waveform to rich contextualized speech representations. We follow the design of HuBERT to build the audio encoder $\textbf{E}_{aud}$, which consists of several temporal convolution networks (TCN) and a transformer encoder. In practical, we utilize the pretrained HuBERT weights as the initialization of our $\textbf{E}_{aud}$, followed by an additional randomly initialized linear projection layer to interpolate the audio features to the target frame rate. The encoded audio features are denoted as  
$A_{1:T}=\{ {a}_{1},\dots,{a}_{T}\in \mathbb{R}^{D}\}$, where $D$ is the feature dimension.

Speaking style is learned from the geometric statistical characteristics of the reference video. In particular, we first use ${C}^{exp}_{1:T}=\{\{0, {\beta}_{1}, 0\}, \dots, \{0, {\beta}_{T}, 0\}\}$ to construct the expression-only mesh vertices ${Vtx}^{exp}_{1:T}=\{{vtx}^{exp}_{1},\dots,{vtx}^{exp}_{T} \in \mathbb{R}^{(3 \times L)} \}$, where $L$ is the total number of vertices. Then, a fully connected layer-based vertex encoder $\textbf{E}_{vtx}$ is introduced to project ${Vtx}^{exp}_{1:T}$ to a lower dimension $D$. Afterwards, the statistical mean $\mu$ and standard deviation $\sigma$ are calculated along the temporal dimension, followed by a fully connected layer to interpolate the concatenation of $\mu$ and $\sigma$ to the style embedding $S \in \mathbb{R}^{D}$. 
We consider it as "style" since it is the overall statistical distribution of facial dynamics. 
It should be noted that, our model learns a universal style representation, and once the model is trained, it can be generalized to any speaker without person-specific fine-tuning as many other approaches \cite{StyleSync, Imitator}.
More specific visual proof can be found in Section \ref{sec:styleembed}.

To inject the style into the audio features, we apply a single cross-attention layer popularized by \cite{attention}, written as
\begin{equation}
    \textit{Attention}(Q, K, V) = Softmax(\frac{Q \times K^\top}{\sqrt{D}}) \times V,
\label{eq:attention}
\end{equation}
where $Q$, $K$ and $V$ refer to the queries, keys and values. Here, we take the style embedding $S$ as $Q$, and take the audio features $A_{1:T}$ as $K$ and $V$.

\noindent\textbf{Lip-Synced Geometry Generation.} 
Firstly, we use shape-only coefficient \{$\overline{\alpha}_{1:T}, 0, 0\}$ to construct the speaker's template mesh vertices, which are then encoded to ${vtx}^{tmp} \in \mathbb{R}^{D}$ by $\textbf{E}_{vtx}$. They are then driven to exhibit synchronized lip movements with the stylized audio features through a cross-attention layer. In particular, the stylized audio features outputted from the first cross-attention layer are set to $Q$, ${vtx}^{tmp} \in \mathbb{R}^{D}$ are designated as $K$ and $V$. Several self-attention layers are followed behind, whose $Q$, $K$ and $V$ are the same values, which are the output feature of the previous layer. Subsequently, a fully connected layer-based vertex decoder $\textbf{D}_{vtx}$ is introduced to project the feature to the target mesh vertices ${Vtx}^{tgt}_{1:T}=\{{vtx}^{tgt}_{1},\dots,{vtx}^{tgt}_{T} \in \mathbb{R}^{(3 \times L)} \}$. Note that we only generate speaker's lower-face vertices which are highly related to the audio, while retain the ground-truth upper-face vertices that provide audio-unrelated expressions such as blinking and frowning. That is, ${vtx}^{tgt}_{t}$ = ${vtx}^{tgt}_{t}[idx_{low}] \cup {vtx}^{gt}_{t}[idx_{up}]$, where $idx_{low}$ and $idx_{up}$ are vertex indexes of lower face and upper face, ${vtx}^{gt}_{t}$ is constructed by $\{{\alpha}_{t}, {\beta}_{t}, 0\}$. Finally, 
we add head pose to ${Vtx}_{1:T}$ and project them to the image plane in the form of PNCC \cite{pncc}, formed as $G_{1:T}$.

\subsection{Dual-Attention Face Rendering}\label{sec:dafr}
In the second stage, we propose a dual-attention face renderer with carefully-designed reference selection strategies to synthesize the target talking faces $F^{tgt}_{1:T}$, supported with the personalized lip-synced geometries $G^{tgt}_{1:T}$, the reference video geometries $G^{ref}_{1:T}$, as well as the reference video frames $F^{ref}_{1:T}$.

\noindent\textbf{Geometry \& Texture Encoding.} 
Considering computational resources and efficiency, we apply attention in the latent space, rather than the pixel space. To this end, we employ a geometry encoder $\textbf{E}_{ref}$ to encode $G^{tgt}_{1:T}=\{ {g}^{tgt}_{1},\dots,{g}^{tgt}_{T} \in \mathbb{R}^{3 \times H \times W}\}$ and $G^{ref}_{1:T}=\{ {g}^{ref}_{1},\dots,{g}^{ref}_{T}\in \mathbb{R}^{3 \times H \times W}\}$ into $\widetilde{G}^{tgt}_{1:T}=\{ \widetilde{g}^{tgt}_{1},\dots,\widetilde{g}^{tgt}_{T}\in \mathbb{R}^{C \times \frac{H}{4} \times \frac{W}{4}}\}$ and $\widetilde{G}^{ref}_{1:T}=\{ \widetilde{g}^{ref}_{1},\dots,\widetilde{g}^{ref}_{T}\in \mathbb{R}^{C \times \frac{H}{4} \times \frac{W}{4}}\}$, and a texture encoder $\textbf{E}_{tex}$ to encode $F^{ref}_{1:T}=\{{f}^{ref}_{1},\dots,{f}^{ref}_{T} \in \mathbb{R}^{3 \times H \times W}\}$ into $\widetilde{F}^{ref}_{1:T}=\{ \widetilde{f}^{ref}_{1},\dots,\widetilde{f}^{ref}_{T}\in \mathbb{R}^{C \times \frac{H}{4} \times \frac{W}{4}}\}$.  Note that these encoders only contain two downsampling convolution layers, which are lightweight yet effective.

\noindent\textbf{Dual-Attention Texture Sampling.} For the target talking face $\widetilde{f}^{tgt}_{t}$ in the latent space, we render its texture using a cross-attention layer described in Equation \ref{eq:attention}. Specifically, we take $\widetilde{g}^{tgt}_{t}$ as $Q$, and select $N$ reference frames including geometries $\widetilde{G}^{ref}_{1:N}$ and textures $\widetilde{F}^{ref}_{1:N}$ as $K$ and $V$. Note that $Q$ is flattened and reshaped from size ($C, \frac{H}{4}, \frac{W}{4}$) to ($\frac{H}{4} \times \frac{W}{4}, C$), $K$ and $V$ are rearranged from size ($N, C, \frac{H}{4}, \frac{W}{4}$) to ($N \times \frac{H}{4} \times \frac{W}{4}, C$). 
where the softmaxed value of the product $Q \times K^\top$ can be seen as the geometry correspondences between $\widetilde{g}^{tgt}_{t}$ and $\widetilde{G}^{ref}_{1:N}$. In this paper, we also add learnable positional embeddings to $Q$ and $K$. The textures of $\widetilde{f}^{tgt}_{t}$ are sampled from reference textures by multiplying geometry correspondences with $\widetilde{F}^{ref}_{1:N}$.
For better preservation of facial details, we separate texture sampling for lip and other facial regions. In particular, we propose a dual-attention structure module consisting of two parallel attention layers, namely $\textbf{Lip-Attention}$ and $\textbf{Face-Attention}$. The former only renders lip with lip-releated geometries and textures. While the latter is responsible for rendering face. These two segments are fused together though a lip mask $M^{l}$. Therefore, the full attention operation becomes 
\begin{equation}
\begin{aligned}
    \textit{Dual-Attention} =\, &\textit{Lip-Attention}(Q^{l}, K^{l}, V^{l}) \times M^{l} \;+ \\
    &\textit{Face-Attention}(Q^{f}, K^{f}, V^{f}) \times (1 - M^{l}).
\end{aligned}
\end{equation}

\begin{algorithm}[t]
\SetAlgoNoLine
\KwIn{Current frame index $i$; Video length $T$; Canonical reference geometries $G^{can}_{1:T}$.}
\KwOut{Face-reference frame indexes $idxs^{f}_{i}$ for frame $i$; Lip-reference frame indexes  $idxs^{l}_{i}$ for frame $i$.}
\SetKwProg{Fn}{Function}{:}{end}
\Fn{Training\_Strategy(i, $T$, $N_{f}$=5, $N_{l}$=5)}{
        $low \Leftarrow Max(1, i - N_{f} * 2)$\;
        $high \Leftarrow Min(T, i + N_{f} * 2)$\;
        $idxs^{f}_{i} \Leftarrow Linspace(low, high, size=n_{f})$\;
        $idxs^{l}_{i} \Leftarrow Random(T, size=N_{l})$\;
        \Return{$idxs^{f}_{i}$, $idxs^{l}_{i}$}
    }

\Fn{Inference\_Strategy(i, $G^{can}_{1:T}$, $N_{f}$=5, $N_{l}$=25)}{
        $low \Leftarrow Max(1, i - N_{f} // 2)$\;
        $high \Leftarrow Min(T, i + N_{f} // 2)$\;
        $idxs^{f}_{i} \Leftarrow Range(low, high, size=N_{f})$\;
        $idxs^{l}\_sorted \Leftarrow Lip\_Argsort(G^{can}_{1:T})$\;
        $idxs^{l}_{i} \Leftarrow \emptyset$\;
        \For{$j$ in $Linspace(1, T, size=N_{l})$}{
        $idxs^{l}_{i} \Leftarrow idxs^{l}_{i} \cup \{idxs^{l}\_sorted_{j}\}$\;
        }
        
        \Return{$idxs^{f}_{i}$, $idxs^{l}_{i}$}
    }

\caption{Reference Selection Strategy}
\label{alg:one}
\end{algorithm}

\noindent\textbf{Reference Frames Selection.} As described above, textures of lip and face are sampled separately, hence we select different references for them. During training, for frame $i$, we uniformly select $N_{f}$ face-reference frames from an \textit{i-centered} frame window with a small window size, and randomly select $N_{l}$ lip-reference frames from the entire video. We argue that selecting the face-reference from adjacent frames, with slight changes in head pose and a static background, could significantly reduce the challenge of texture sampling and force the network to learn subtle facial muscle dynamics, thereby improving the facial stability and fidelity. During inference, for every \textit{i-th} frame, we directly take \textit{i-centered} $N_{f}$ frames as the face-reference, since head poses and lighting in these frames are most similar to the current frame, allowing for better preservation. As for lip-reference, we utilize the canonical geometries $G^{can}_{1:T}$, constructed by ${C}^{can}_{1:T}=\{\{{\alpha}_{1}, {\beta}_{1}, 0\}, \dots, \{{\alpha}_{T}, {\beta}_{T}, 0\}\}$, to perform a global sorting of mouth opening-size, and uniformly select the reference from these sorted frames. It should be noted that, attention mechanism naturally extends to any length of $K$ and $V$, thus $N_{l}$ selected for inference can be much greater than that for training, e.g., 5x or more. It ensures both diversity and global consistency of lip-reference texture, effectively capturing nuanced lip dynamics, improving lip fidelity, and eliminating common tooth flicker and sticking artifacts. We set $N_{f}$ = $N_{l}$ = 5 during training, and $N_{f}$ = 5, $N_{l}$ = 25 during inference. Please refer to Algorithm~\ref{alg:one} for the pseudo code.

\noindent\textbf{Texture Decoding.} After texture sampling, we obtain the final talking faces $\widetilde{F}^{tgt}_{1:T}=\{ \widetilde{f}^{tgt}_{1},\dots,\widetilde{f}^{tgt}_{T}\in \mathbb{R}^{(\frac{H}{4} \times \frac{W}{4}) \times C}\}$ in the latent space. We first restore their spatial structure, specifically rearranging them from size ($\frac{H}{4} \times \frac{W}{4}, C$) back to ($C, \frac{H}{4}, \frac{W}{4}$). After that, we adopt a geometry-aware texture decoder $\textbf{D}_{tex}$ to decode them to the pixel space. Similar to the encoders, the texture decoder only includes two upsampling convolution layers. Note that to protect the facial geometry during decoding, we interpolate $g^{tgt}$ to the same shape as $\widetilde{f}^{tgt}$ and concatenate them in the channel dimension before sending them into the $\textbf{D}_{tex}$.

\section{expriments}

\begin{figure*}
  \centering
  \includegraphics[width=0.9\linewidth]{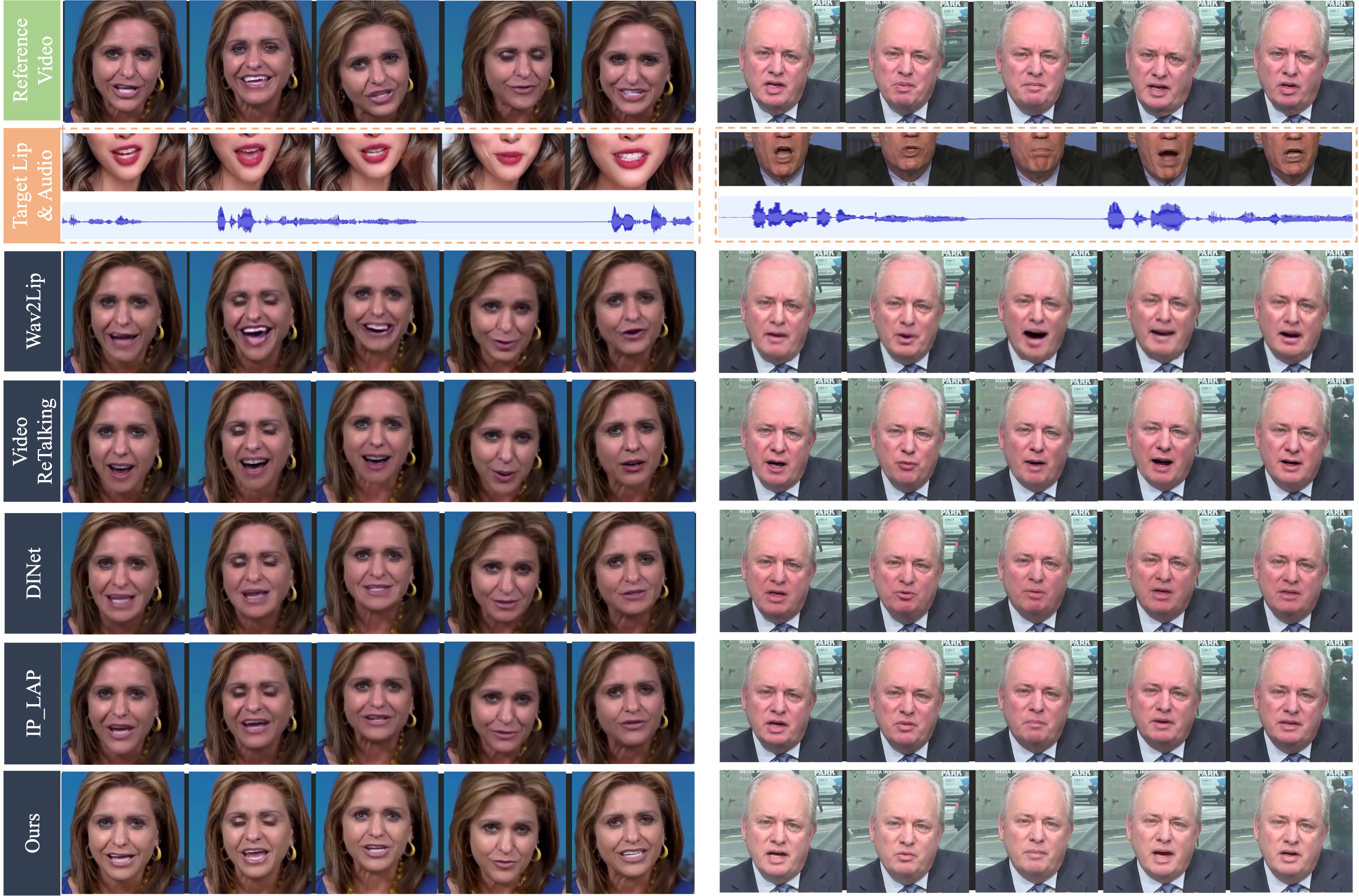}
  \caption{Qualitative comparisons with Wav2Lip \cite{Wav2Lip}, VideoRetalking \cite{videoretalking}, DINet \cite{DINet} and IP\_LAP \cite{IP-LAP}. The top row is the input (reference) video, the second row is target lip movements with target audio. Our method not only generates accurate lip movements, but also preserves speaker's speaking style and facial details.}
  \label{qualitative}
\end{figure*}

\subsection{Experimental Settings}

\noindent\textbf{Datasets.}
We utilize a mixture of datasets, including VoxCeleb2 \cite{VoxCeleb2}, CelebV-HQ \cite{CelebV-HQ}, and VFHQ \cite{VFHQ} to train our $PersonaTalk$.
VoxCeleb2 is a large-scale audio-visual dataset that contains over 1 million utterances for 6,112 celebrities, extracted from videos uploaded to YouTube.
CelebV-HQ is a large-scale, high-quality, and diverse video dataset consists of 35,666 video clips involving 15,653 identities with rich facial attribute annotations.
VFHQ is a high-quality video face dataset for super-resolution, which contains over 16,000 high-fidelity clips of diverse interview scenarios.
We apply the state-of-the-art Image Quality Assessment (IQA) method \cite{su2020blindly} to filter out videos, which are of low quality.

For a fair comparison, we choose an out-of-domain dataset HDTF \cite{HDTF}, which comprises about 362 different videos for 15.8 hours with original resolution 720P or 1080P, as our evaluation dataset. 30 videos in HDTF are randomly selected for quantitative comparison.

\noindent\textbf{Implementation Details.}
In our experiments, videos are first converted to 25 fps and then cropped to the size of 256 $\times$ 256, according to face landmarks extracted by the MediaPipe \cite{MediaPipe}. Audios are resampled to 16kHz.

\subsection{Evaluation Metrics}
The experimental results are evaluated in three aspects: visual quality, lip-sync accuracy and persona preservation.

{\noindent\textbf{Visual Quality.}
Following previous studies \cite{Wav2Lip,StyleSync,IP-LAP}, we use Structured Similarity (SSIM), Peak Signal-to-Noise Ratio (PSNR) and Frechet Inception Distance (FID) to evaluate the visual quality.

{\noindent\textbf{Lip-Sync Accuracy.}
In terms of lip-sync accuracy, we utilize normalized landmark distance around the lip (LMD) \cite{StyleSync}, and the confidence score of SyncNet (SyncScore) \cite{SyncNet} . 

{\noindent\textbf{Persona Preservation.}
We compute the cosine similarity (CSIM) of identity vectors extracted by the face recognition model ArcFace \cite{ArcFace} to evaluate the identity preservation. LPIPS \cite{LPIPS} is employed to measure the feature-level similarity between generated and ground-truth faces. 

Currently, there is no suitable evaluation metric for speaking style preservation in visual dubbing. In this paper, we introduce the Speaking Style Similarity (StyleSim). Given two videos $V_1$ and $V_2$, we first extract their expression-only vertices $Vtx^{exp}_1$ and $Vtx^{exp}_2$ as described in Section \ref{sec:sagc}, following by mapping them to style representations $S_{1}$ and $S_{2}$ with a pretrained speaking style encoder $\Phi$. Note that we refer to \cite{diffposetalk} for the design of its model structure. Ultimately, we consider the cosine similarity of $S_{1}$ and $S_{2}$ as the StyleSim.

\subsection{Comparisons}
{\noindent\textbf{Comparison Methods.}
We compare our method against SOTA person-generic methods, including Wav2Lip \cite{Wav2Lip}, VideoRetalking \cite{videoretalking}, DINet \cite{DINet} and IP\_LAP \cite{IP-LAP}. 
Wav2Lip trains the model with a pretrained lip-expert, to improve the audio-visual synchronization.
VideoRetalking introduces delicate designs of pre-/post-processing on generating photo-realistic talking faces.
DINet performs spatial deformation on feature maps to preserve high frequency details, yielding high-fidelity results.
{IP\_LAP} leverages landmark and appearance prior information to preserve identity.
In addition, we also compare with the person-specific approach SyncTalk \cite{SyncTalk}, a NeRF based method which demonstrates state-of-the-art performance.

{\noindent\textbf{Quantitative Comparisons.}
As shown in Table \ref{tab:1}, our method achieves the best visual qualities according to SSIM, PSNR and FID. As for the lip-sync accuracy, our method still gets much better and comparable performance than other approaches. Of particular note, in terms of persona preservation, our method is far ahead of other approaches. This reveals the progressiveness of our proposed style-aware geometry construction module and dual-attention face rendering module in capturing speaker's speaking style and upholding speaker's facial details.

Table \ref{tab:2} shows that our method can achieve comparable performance as the SOTA person-specific method without any further fine-tuning.

\begin{table*}[t]
\caption{Quantitative Comparisons with Person-Generic Methods}
\centering
\begin{tabular}{l|ccc|cc|cccc}
\toprule
& \multicolumn{3}{|c|}{Visual Quality} & \multicolumn{2}{c|}{Lip-Sync} & \multicolumn{3}{c}{Persona} \\
    & SSIM $\uparrow$ & PSNR $\uparrow$ & FID $\downarrow$ & LMD $\downarrow$ & SyncScore $\uparrow$ & LPIPS $\downarrow$ & CSIM $\uparrow$ & StyleSim $\uparrow$ \\
\hline

Wav2Lip \cite{Wav2Lip} & 0.934 & 28.347 & 31.561 & 0.376 & \textbf{7.147} & 0.092 & 0.948 & 0.896 \\
VideoReTalking \cite{videoretalking} & 0.911 & 26.963 & 35.025 & 0.573 & 6.615 & 0.053 & 0.931 & 0.846 \\
DINet \cite{DINet} & 0.914 & 26.506 & 30.355 & 0.657 & 5.771 & 0.057 & 0.896 & 0.865 \\
IP\_LAP \cite{IP-LAP} & 0.951 & 30.717 & 20.549 & 0.405 & 4.409 & 0.044 & 0.947 & 0.892 \\
\hline

Ours &   \textbf{0.956} & \textbf{31.451} & \textbf{10.701} & \textbf{0.313} & 6.610 & \textbf{0.018} & \textbf{0.983} & \textbf{0.939}  \\
GT & 1.000 & N/A  & 0.000 & 0.000 & 6.134 & 0.000 & 0.988 & 0.941 \\
\bottomrule
\end{tabular}

\vspace{-0.1cm}
\label{tab:1}
\end{table*}


{\noindent\textbf{Qualitative Comparisons.}
Subject evaluation is crucial for determining the quality of the results for visual dubbing. Here we show the comparison of our methods against SOTA methods on two cases in Figure \ref{qualitative}. It can be seen that, Wav2Lip has poor visual quality with artifacts and blurry. VideoRetalking and DINet fail to capture speaker's speaking style, leading to the homogenized lip movements. IP\_LAP cannot always generate lip-synced results. Besides, all of them are unable to preserve the details of the speaker's teeth. To this end, our method achieves the best generation quality and person preservation.

Additionally, we apply self-dubbing comparison with SOTA person-specific method SyncTalk and show the result in Figure \ref{fig:synctalk}. As shown in the figure, our method generates competitive results in terms of visual quality, lip-sync accuracy and persona preservation. More qualitative results are shown in the supplementary video.

{\noindent\textbf{User Study.} 
To evaluate the performance of our proposed model, we conduct a Mean Opinion Score (MOS) evaluation involving four other models alongside our own model. Each model generates 10 sentences, resulting in 60 sentences which are then randomly shuffled. Human evaluators rate these sentences on a scale of 1 (worst) to 5 (best) based on three dimensions: persona preservation, lip-sync accuracy and visual quality. Results are shown in Table \ref{tab:userstudy}. As can be seen, our method outperforms its counterparts in all the three aspects by a large margin.

\begin{figure}
    \centering
    \includegraphics[width=\linewidth]{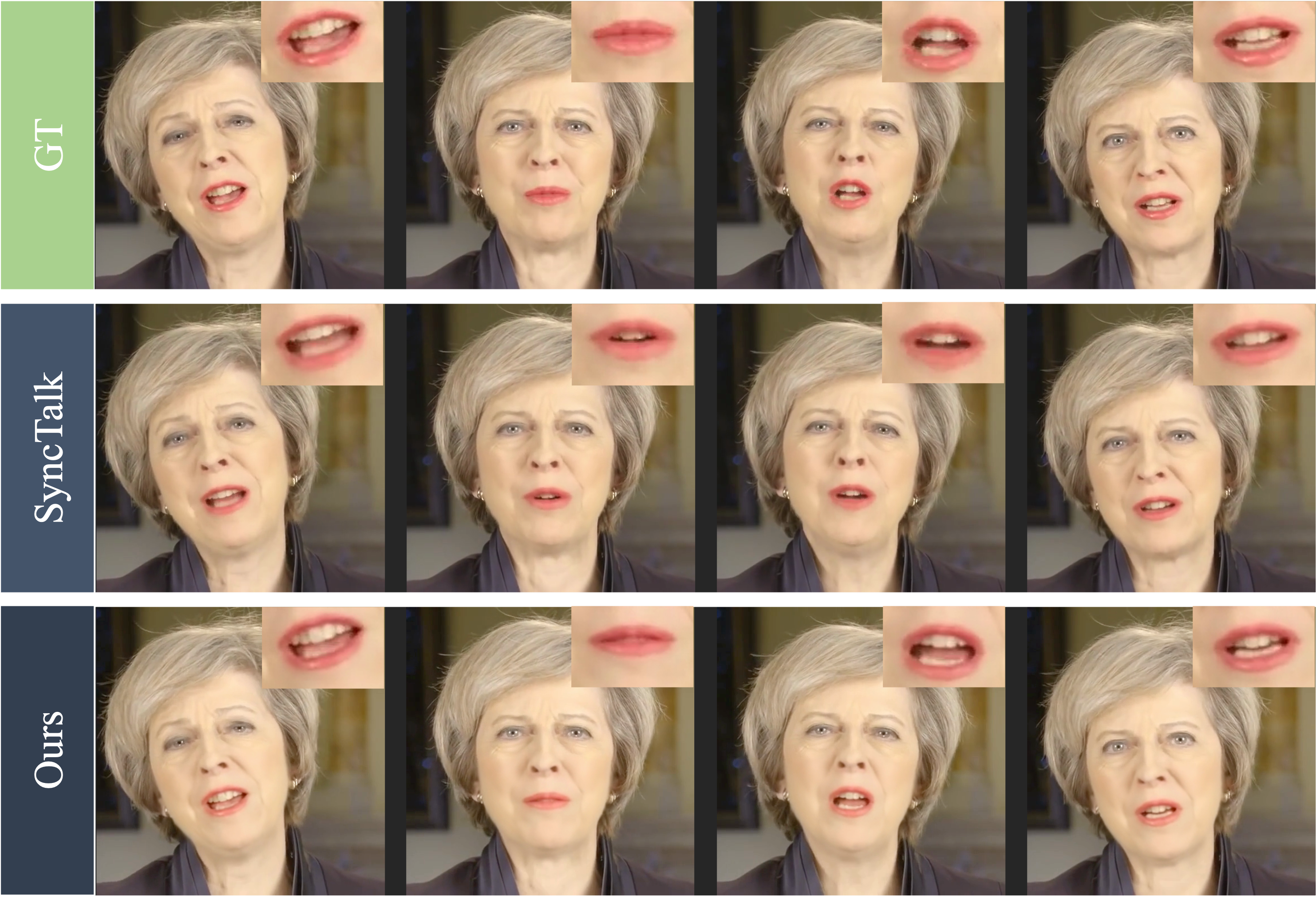}
    \caption{Self-dubbing results compared with SOTA person-specific method.}
    \label{fig:synctalk}
\end{figure}

\begin{table}[t]
\caption{Quantitative Comparisons with SOTA Person-Specific Method}
\centering
\begin{tabular}{lccccc}
\toprule
& SyncScore $\uparrow$ & LPIPS $\downarrow$ & CSIM $\uparrow$ & StyleSim $\uparrow$ \\ \hline
SyncTalk & \textbf{8.070} &  \textbf{0.013} &  \textbf{0.970} &  0.916 \\
Ours & 7.703 & \textbf{0.013}  & \textbf{0.971} & \textbf{0.927} \\
\bottomrule
\end{tabular}
\vspace{-8pt}
\label{tab:2}
\end{table}


\subsection{Ablation Study}
We ablate two major modules of our framework in Table \ref{tab:ablation}. Specifically, we remove the style encoding in the first stage. As we expected, the generated results cannot preserve the speaker's speaking style, reflected by the decrease of the StyleSim.
Besides, we change the dual-attention structure to a single cross-attention layer. We observe that there is degradation in the quality of lip and teeth generation as well as the stability of the face outline. Correspondingly, the LPIPS increases while the CSIM decreases. Finally, we forgo the frame selection strategy and randomly select reference frames for face and lip. Results indicate that the generation ability of the model has severely degraded.

\begin{figure}
    \centering
    \includegraphics[width=\linewidth]{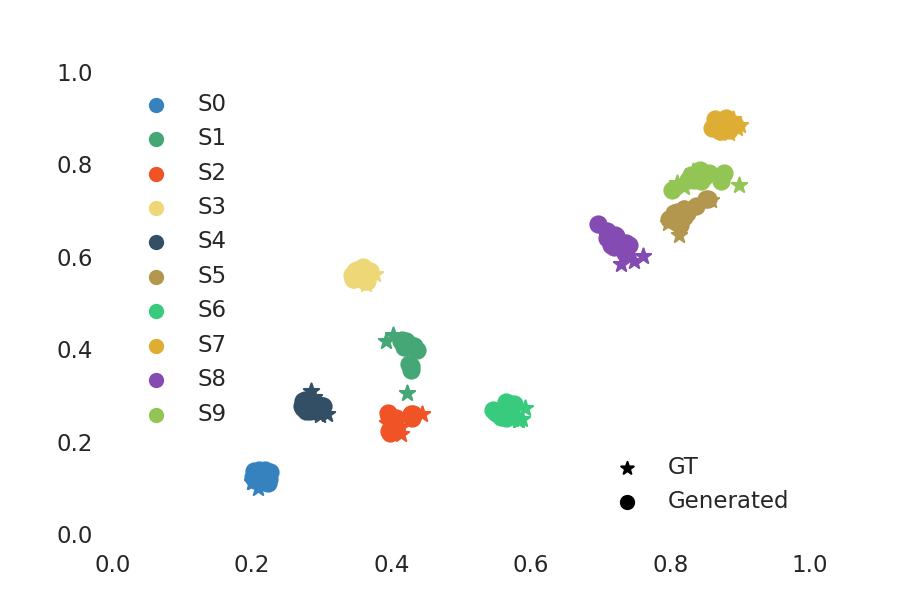}
    \caption{The t-SNE visualization of our extracted style embeddings. Points of different colors indicate different speakers. "GT" implies original videos and "Generated" represent data generated by our method.}
    \label{fig:tsne}
\end{figure}

\subsection{Visualization of Style Embeddings.} \label{sec:styleembed}
We first randomly select 10 speakers in the test set and generate 12 dubbing videos for each speaker using our method. Then we extract the style embeddings from these videos as well as the original videos with our style encoding module. We visualize these embeddings in Figure \ref{fig:tsne}, which are dimension-reduced by t-SNE \cite{t-SNE}. As can be seen, style embeddings of different speakers are well-clustered. Additionally, style embeddings extracted from the generated dubbing videos are closely grouped with the embeddings extracted from the original video, demonstrating our model's ability to preserve speaker's speaking style in visual dubbing.

\begin{table}[t]
\caption{User Study Measured by Mean Opinion Scores}
\centering
\begin{tabular}{lccccc}
\toprule
       & Visual & Lip-Sync & Persona \\ \hline
Wav2Lip & 1.478 &  2.261 &  2.112 \\
VideoReTalking & 2.462 & 2.346  & 2.115 \\
DINet & 2.227 & 3.045  & 2.818 & \\
IP\_LAP & 2.511 & 2.077  & 2.962 & \\
\midrule
Ours & \textbf{4.739} & \textbf{4.303}  & \textbf{4.870} & \\
\bottomrule
\end{tabular}
\vspace{-8pt}
\label{tab:userstudy}
\end{table}

\begin{table}[t]
\caption{Ablation Studies for Persona Preservation}
\centering
\begin{tabular}{lcccccc}
\toprule
       & LPIPS $\downarrow$ &  CSIM $\uparrow$ & StyleSim $\uparrow$ \\ \hline
w/o style embedding     &  0.024  &  0.981  &  0.899  \\
w/o face-attention      &  0.029  &  0.976  &  0.937   \\
w/o lip-attention       &  0.030  &  0.971  &  0.921   \\
w/o frame-selection     &  0.038  &  0.966  &  0.901   \\
\midrule
Ours  & \textbf{0.018}  & \textbf{0.983} & \textbf{0.939} \\
\bottomrule
\end{tabular}
\label{tab:ablation}
\end{table}

\section{conclusion}
In this paper, we propose an attention-based two-stage framework to achieve high-fidelity and persona-preserved visual dubbing. As demonstrated, our method could capture video speaker's unique speaking style, and uphold speaker's intricate facial details.
Extensive experiments and user studies demonstrate our advantages over other state-of-the-art methods. Moreover, as a person-generic framework, our approach can achieve competitive performance as person-specific methods.

\noindent\textbf{Limitations.}
Due to the limited diversity of our training data, our model's performance in driving non-human avatars, such as cartoons, may exhibit slightly lower effectiveness. Furthermore, artifacts may occur in facial generation with large face postures.

{\noindent\textbf{Ethical Considerations}. We recognize the potential misuse of our method to fabricate deceptive talks and speeches. As a precaution, we will restrict access to our core models exclusively to research institutions. Additionally, we want to clarify that our options for portrait video examples are restricted by the necessity to conduct comparisons with existing works. These portraits are all from publicly available datasets.

\bibliographystyle{ACM-Reference-Format}
\bibliography{bibliography}

\end{document}